Title of Abstract:
**Automatic Annotation of Axoplasmic Reticula in Pursuit of Connectomes using High-Resolution Neural EM Data**

Authors' Names and Affiliations:
Ayushi Sinha[†], William Gray Roncal[†‡], Narayanan Kasthuri[§¤], Jeff W. Lichtman[§¤], Randal Burns[†], Michael Kazhdan[†]
[†]Department of Computer Science, The Johns Hopkins University, Baltimore, MD
[‡]The Johns Hopkins University Applied Physics Laboratory, Laurel, MD
[§]Department of Molecular and Cellular Biology, Harvard University, Cambridge, MA
[¤]Center for Brain Science, Harvard University, Cambridge, MA

Conferences or Organizations That Have Published This Abstract: N/A

Synopsis (100 words or less):
Accurately estimating the wiring diagram of a brain, known as a connectome, at an ultrastructure level is an open research problem. Specifically, precisely tracking neural processes is difficult, especially across many image slices. Here, we propose a novel method to automatically identify and annotate small subcellular structures present in axons, known as axoplasmic reticula, through a 3D volume of high-resolution neural electron microscopy data. Our method produces high precision annotations, which can help improve automatic segmentation by using our results as seeds for segmentation, and as cues to aid segment merging.

Declaration of Speaker Financial Interests or Relationships: N/A

Categories:
    Acquisition (modality used): Electron Microscopy
    Analysis: Computer Vision, Connectome
    Application:
        Organ System: N/A
        Subjects (humans, animals, cells): Mouse, Human
        Physiology/Pathology: N/A

Notes: N/A

**Title**: Automatic Annotation of Axoplasmic Reticula in Pursuit of Connectomes using High-Resolution Neural EM Data

**Introduction**: The Open Connectome Project [1] facilitates the storage and analysis of image and annotation data to ultimately estimate wiring diagrams of the brain, known as connectomes. This problem is challenging because there are approximately 100 billion neurons, 100 trillion synapses, and exabytes of data in a single human brain. These connectomes will help elucidate the structure and function of the brain, and potentially lead to advances in understanding and curing disease, learning, and technology. This work presents a novel, highly precise method to detect axoplasmic reticula (AR) present in axonal processes. Our results can be used to support scientific inquiry, identify axon locations, and enable the improved segmentation and tracking necessary to form structural connectomes.

**Subjects and Methods**: We developed our algorithm on the Kasthuri11 dataset [2], a high-resolution volume (i.e., 3x3x30nm anisotropic voxels) of neural electron microscopy (EM) data, which allows for the analysis of very small subcellular structures such as AR, vesicles, and mitochondria. Each slice in this 3D volume initially contained high contrast variations. We color corrected the data using gradient-domain image-stitching techniques [3] to adjust contrast through the slices.

We then filtered this color corrected data using an edge preserving non-linear smoothing filter, known as the bilateral filter. It consists of two Gaussian kernels, one which decays with distance and the other with intensity [4], hence smoothing out noise in the data while preserving important detail such as edges. Although this operation accentuates AR in our data, it also causes some amount of color bleeding across edges even with a narrow Gaussian in the intensity domain. We mitigated this effect through Laplacian sharpening, which adds to the original image the difference in intensity between a pixel and its neighbors, hence highlighting edges around AR.

Next, we used a region growing technique on the filtered data to identify AR. Our method iterates over each slice looking for candidate AR centers based on intensity values, and grows outward from each center by iteratively checking its neighborhood to locate AR based on biological priors (e.g., shape, size, color). We run this algorithm on the bilateral filtered data, as well as on the bilateral and Laplacian filtered data.

Once we have annotations on each slice, we perform an automated error check to eliminate false positives. We know that AR are contiguous structures within axons; therefore, we require an annotated AR to persist on at least one adjacent slice. We enforce this criteria using a less restrictive growing algorithm, and use the results to remove spurious annotations and include new detections.

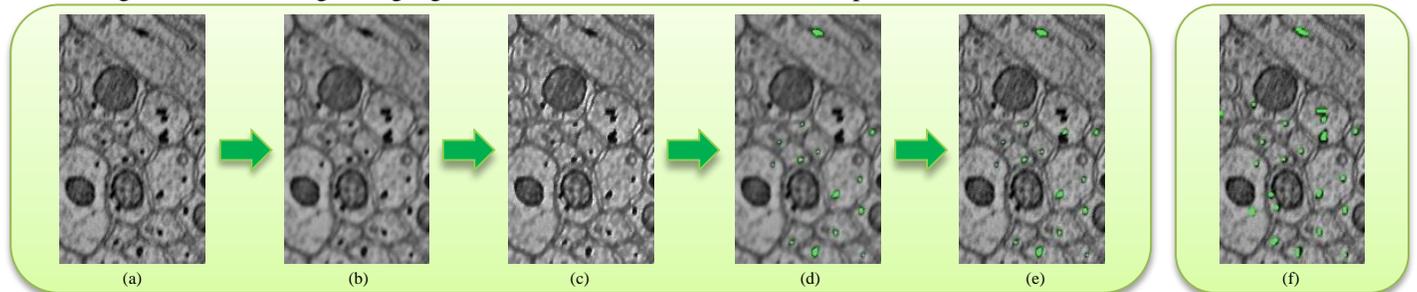

Figure 1: (a) Color corrected data, (b) bilaterally filtered data, (c) Laplacian filtered, (d) annotations from the region growing algorithm, (e) annotations after error checking, and (f) ground truth

**Results and Discussion**: We qualitatively evaluated our annotations on 20 slices from the Kasthuri11 dataset [2], and quantitatively compared our results against ground truth provided by a neurobiologist. Our algorithm is able to annotate AR with 87% precision and 52% recall, which is sufficient to derive contextual information, and provides a baseline for the community. Note that there is inherent ambiguity in labeling even among expert annotators. We are currently using our results as input for tracking algorithms based on directional displacement, and plan to incorporate more robust techniques such as Kalman and particle filtering. Finally, our method can be adapted to annotate similar features (e.g., mitochondria) by modifying thresholds.

**Conclusion**: Our method to find AR achieves high precision, which enables contextual inference, improved segmentation methods, and ultimately, highly accurate estimates of neural connectivity.

**Funding**: This work was supported in part by grant number 1RO1EB016411-01 from the National Institute of Biomedical Imaging at the National Institutes of Health.